    \documentclass{article}
    \usepackage{spconf,amsmath,graphicx,hyperref}
    \usepackage{booktabs}
    \usepackage{multirow}
    \usepackage{array}
    \usepackage{tikz}
    \usetikzlibrary{tikzmark}
    
    \title{Benchmarking Answer-Grounding Consistency in Remote Sensing Visual Question Answering}
\name{
    Ping Guo$^{1}$,
    Zhiqi Huang$^{2}$,
    Xinran Li$^{3}$,
    Linkun Yu$^{4}$,
    Jingyin Pei$^{1}$,
     Kamata Seiichiro$^{1}$\sthanks{Corresponding author.}
    }
    
    \address{
    $^{1}$ Graduate School of Information, Production and Systems, Waseda University, Kitakyushu, Japan \\
    $^{2}$ School of Artificial Intelligence, The Chinese University of Hong Kong, Shenzhen, China \\
    $^{3}$ School of Software, Dalian University of Technology, Dalian, China \\
    $^{4}$ Graduate School of Fundamental Science and Engineering, Waseda University, Tokyo, Japan \\
    {\small The code is available at \url{https://github.com/pingggg516/AESP-Evaluation}.}
}

\begin{document}

    %
    \maketitle
    \begin{abstract}
    In grounded remote sensing visual question answering, a model produces a textual answer and identifies relevant objects in the image as grounding evidence. Existing evaluations measure answer correctness and localization quality, but do not directly assess answer-grounding consistency. We propose the Answer-Evidence Support Protocol (AESP) to assess this consistency through evidence support: whether the grounding evidence supports the model's own answer. AESP applies task-specific rules to existence, counting, and comparison questions. We further construct the Grounded Remote Sensing Visual Question Answering (G-RSVQA) benchmark, containing 15,027 samples with questions, reference answers, and all related object-level annotations. Experiments on seven grounded vision--language models reveal failures of evidence support across all three question types. Up to 50.82\% of correct answers in the evidence evaluation subset are not supported by the provided grounding evidence.
    \end{abstract}
    \begin{keywords}
    Remote sensing visual question answer, visual grounding, answer-grounding consistency, benchmark
    \end{keywords}

    \section{Introduction}
    \label{sec:intro}
    
    Remote sensing imagery provides large-scale observations of the Earth's surface, but obtaining task-specific information from rapidly growing image collections often requires expert knowledge. Remote sensing visual question answering (RS-VQA) provides a natural-language interface for querying image content~\cite{lobry2020rsvqa}. Grounded RS-VQA further produces object localizations together with textual answers~\cite{kuckreja2024geochat}. These grounding regions can serve as visual evidence, allowing users to check an answer against the image. However, accurate localization of relevant objects does not guarantee these objects collectively support the claim stated in the textual answer.

    \begin{figure}[t]
        \centering
        \includegraphics[width=1.0\columnwidth]{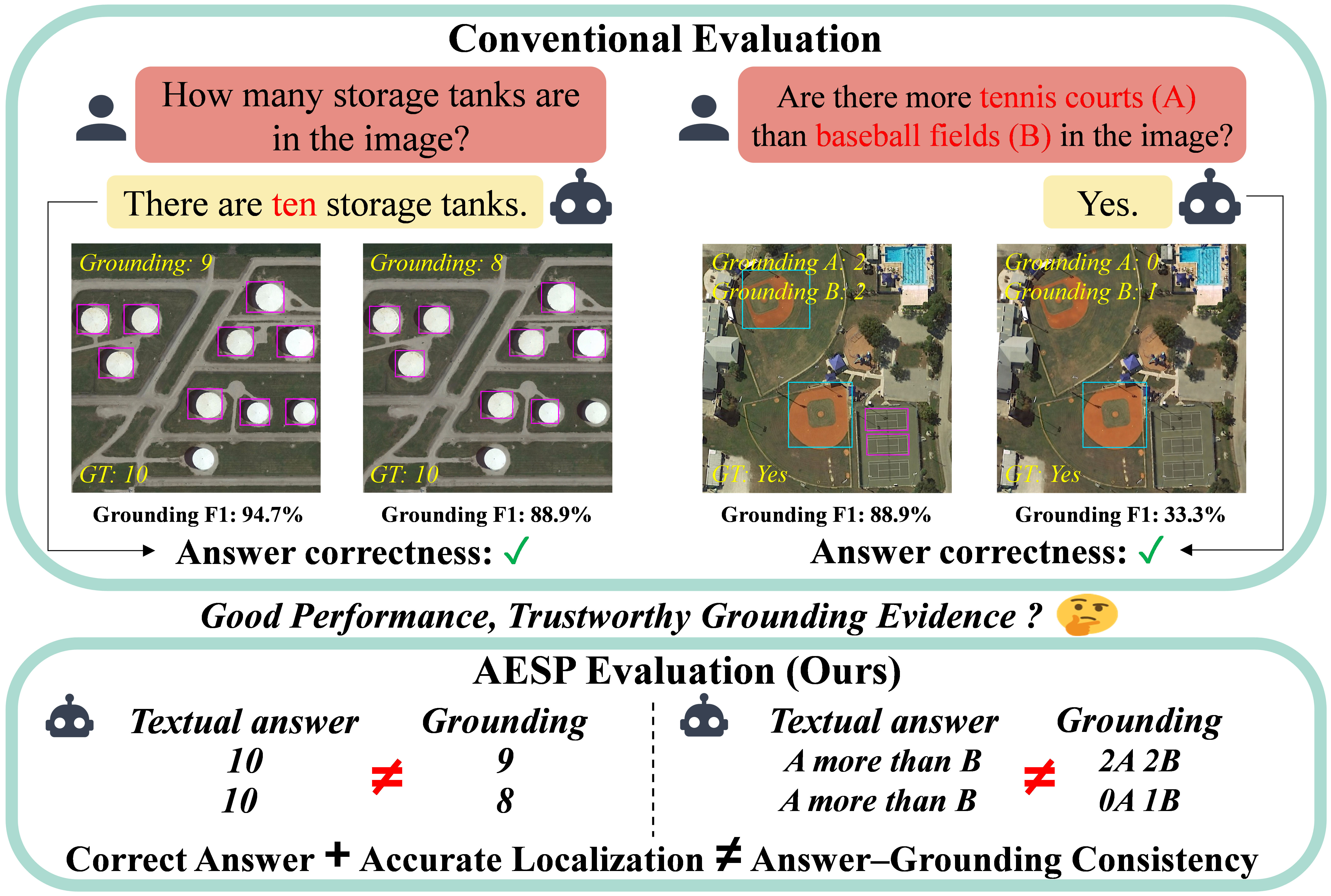}
        \par\vspace{-8pt}
        \caption{Conventional evaluation measures answer correctness and localization quality; AESP additionally assesses answer-grounding consistency.}
        \label{fig:motivation}
    \end{figure}
    
    Existing remote sensing vision--language research has expanded beyond predicting textual answers to include visual grounding. RSVQA~\cite{lobry2020rsvqa} provides benchmarks for question answering, while GeoChat~\cite{kuckreja2024geochat} supports both VQA and grounded conversations that link textual responses to object coordinates. GeoGround~\cite{zhou2024geoground} further broadens grounding capabilities by unifying bounding-box and segmentation-mask outputs. VRSBench~\cite{li2024vrsbench} brings question answering, visual grounding, and image captioning into a common benchmark. Despite this broader task coverage, GeoChat and VRSBench evaluate VQA and grounding as separate tasks. They report answer accuracy for VQA and assess localization by comparing predicted boxes with ground-truth boxes using intersection-over-union (IoU) thresholds.

    \begin{figure*}[t]
        \centering
        \includegraphics[width=\textwidth]{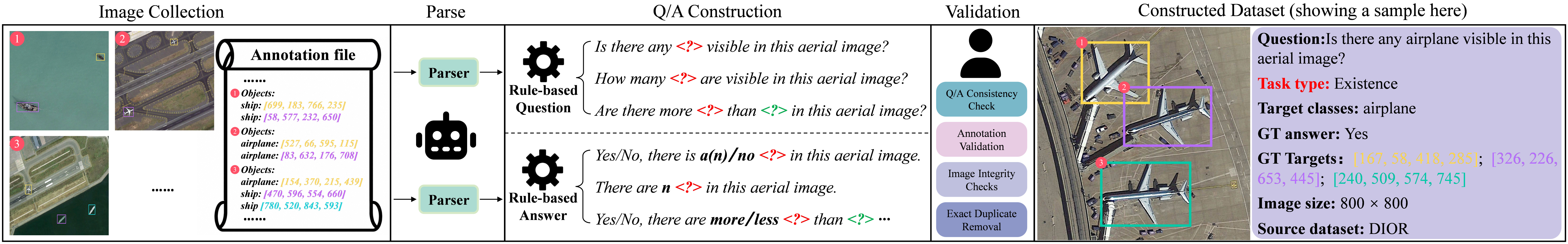}
        \par\vspace{-8pt}
        \caption{Overview of G-RSVQA construction. The rightmost panel shows a sample and its annotation format.}
        \label{fig:Process}
    \end{figure*}
    
    Beyond remote sensing, studies address feature--threshold calibration~\cite{guo2026noise}, cross-teacher consistency~\cite{guo2026ctfs}, and joint evaluation. NExT-GQA~\cite{xiao2024can} evaluates answer correctness and the localization of relevant video segments. MM-GCoT~\cite{wu2026grounded} measures consistency using answer and bounding-box correctness, while VISTAQA~\cite{azadani2026vistaqa} combines answer correctness and mask quality into a joint score. For counting and comparison questions, however, combining or jointly evaluating these scores can not explicitly check whether the grounded object counts agree with the count or numerical relation stated in the answer. As illustrated in Fig.~\ref{fig:motivation}, grounding nine of ten storage tanks yields a high localization F1-score but does not support the textual answer \textbf{``ten''}. Similarly, grounding equal object numbers of two  classes can not support the answer claiming one class has more instances than the other.
    
    To address this gap, we propose the Answer-Evidence Support Protocol (AESP) to assess answer-grounding consistency. Using object-level annotations, AESP counts instances covered by predicted grounding regions and applies task-specific rules to assess support for the model's own answer. Answer correctness and localization quality are evaluated separately. We further construct the Grounded Remote Sensing Visual Question Answering (G-RSVQA) benchmark with all related object-level annotations.
    
    \begingroup
    \setlength{\parskip}{0pt}
    \setlength{\parindent}{1em}
    
    Our main contributions are as follows:\par
    \textbf{1)} We propose AESP with task-specific rules for existence, counting, and comparison questions to assess whether grounding evidence supports the model's own answer.\par
    \textbf{2)} We construct G-RSVQA with 15,027 question--answer samples and all related object-level annotations.\par
    \textbf{3)} We evaluate seven grounded vision--language models, finding that up to 50.82\% of correct answers in the evidence evaluation subset lack supporting evidence.\par
    
    \endgroup

    \section{G-RSVQA Benchmark}

    We construct G-RSVQA to evaluate evidence support for existence, counting, and comparison questions. This evaluation requires verifying which objects the predicted grounding regions identify and whether the numbers of grounded instances support the model's answer. Each question is therefore paired with a reference answer and all related object-level annotations, covering every instance of the target classes so that any objects selected by a model as evidence can be verified.
    
    As shown in Fig.~\ref{fig:Process}, we collect images and object annotations from DIOR~\cite{li2020object}, NWPU VHR-10~\cite{cheng2014multi}, and SODA-A~\cite{cheng2023towards} datasets. We parse the annotations to extract object classes and locations, then generate questions and reference answers using rules based on object presence, instance counts, and numerical relations between classes. For each question, we retain all related object-level annotations from which its reference answer is derived. We check question--answer consistency, annotation validity, and image integrity, and remove exact duplicates. The rightmost panel shows a constructed sample and its annotation format.

    \begin{figure}[t]
        \centering
        \begin{minipage}[b]{0.41\linewidth}
            \centering
            \includegraphics[height=3.6cm]{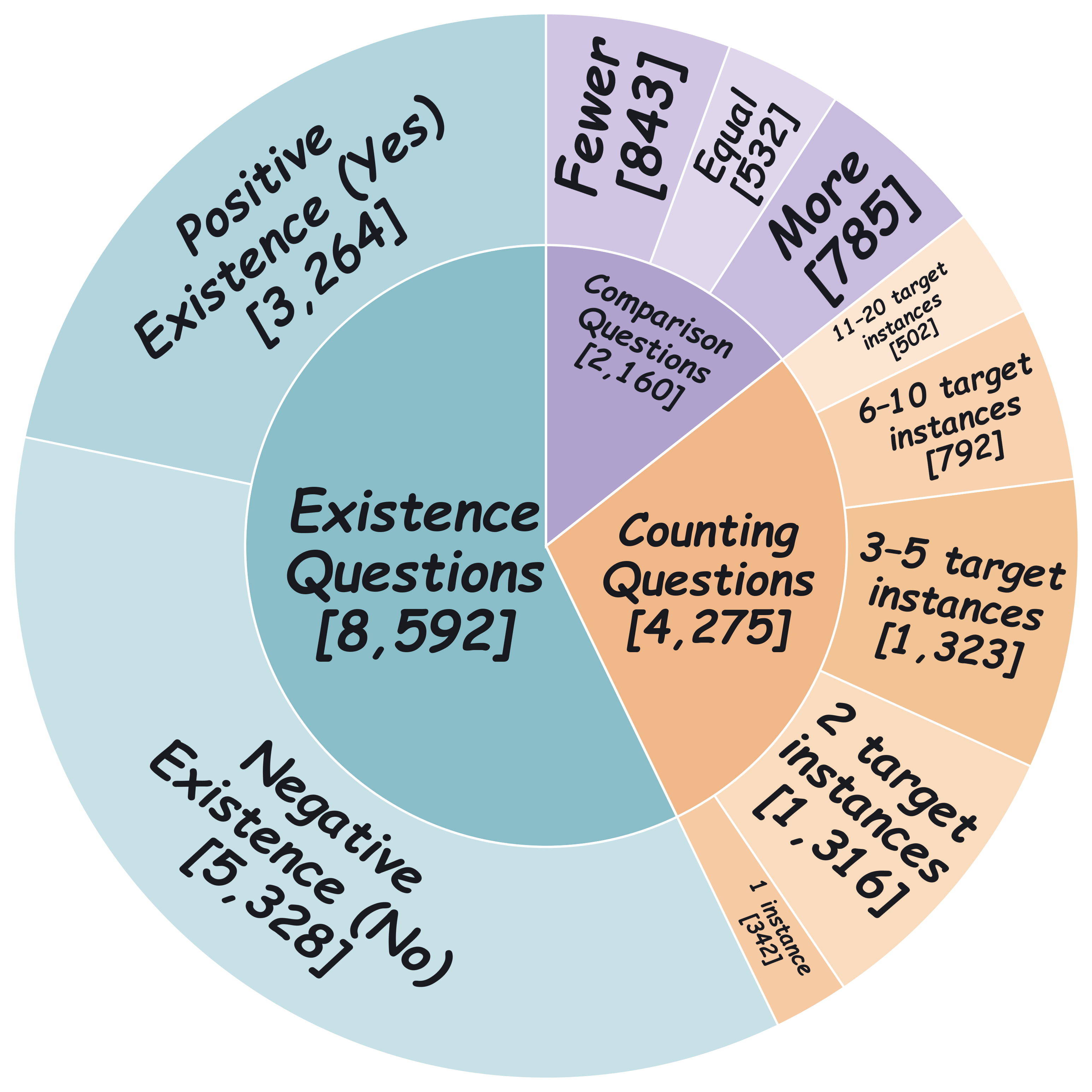}
            \par\vspace{-2pt}
            \makebox[\linewidth][c]{(a) Question composition}
        \end{minipage}
        \hfill
        \begin{minipage}[b]{0.58\linewidth}
            \centering
            \includegraphics[height=3.6cm]{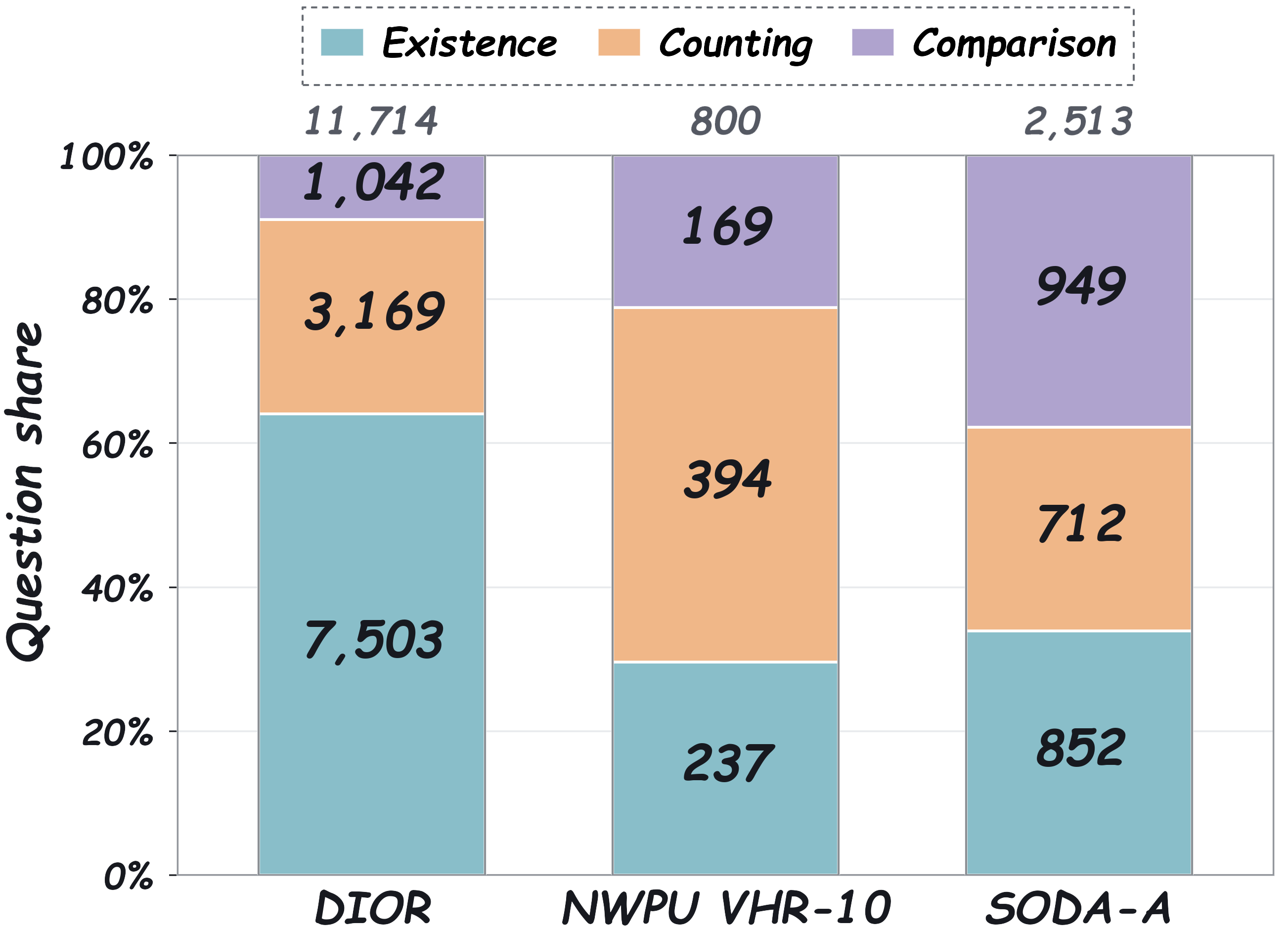}
            \par\vspace{-2pt}
            \makebox[\linewidth][c]{\hspace*{6mm}(b) Dataset sources}
        \end{minipage}
        \par\vspace{-7pt}
        \caption{Question types and data sources of G-RSVQA.}
        \label{fig:benchmark_introduction}
    \end{figure}

    As shown in Fig.~\ref{fig:benchmark_introduction}, G-RSVQA contains 15,027 question--answer samples: 8,592 existence, 4,275 counting, and 2,160 comparison questions. These cover object presence and absence, counts from zero to more than ten target instances, and fewer-than, equal-to, and more-than relations between classes. Each dataset includes all three question types.

    \section{AESP Evaluation Protocol}
    \label{sec:AESP}

    \subsection{Answer Extraction and Grounding Verification}
    
    We normalize model responses to Yes/No answers for existence and comparison questions and nonnegative integers for counting questions. Responses already in these formats are parsed directly; otherwise, Qwen3-14B~\cite{yang2025qwen3} extracts the answer using only the question and original response. Ambiguous or contradictory responses are invalid. After converting predicted regions to original image coordinates, we count covered target instances per class using all related object-level annotations. Each instance is counted once if its ground-truth box center lies within any predicted region.

    \begin{figure*}[t]
        \centering
        \includegraphics[width=0.98\textwidth]{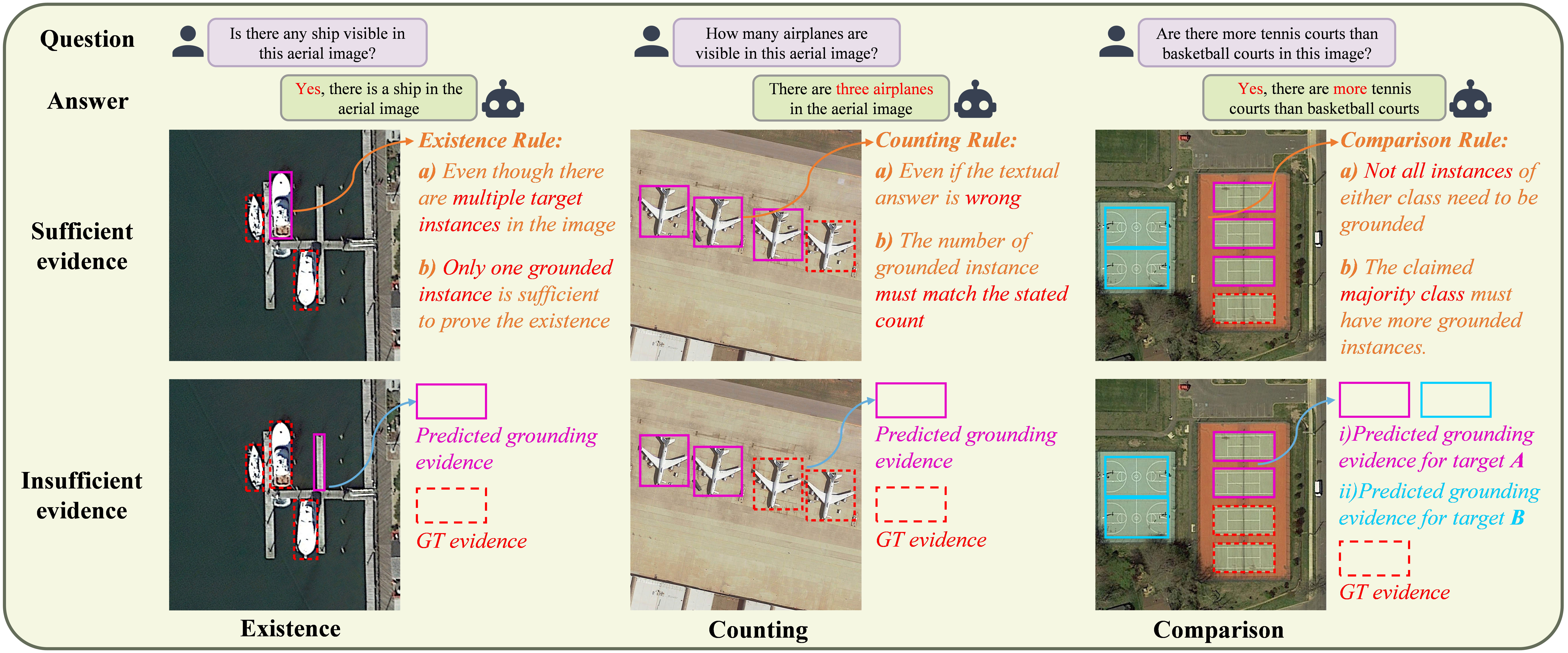}
        \par\vspace{-6pt}
        \caption{Illustration of AESP evaluation protocol for assessing whether grounding evidence supports a model’s answer. Each question is paired with supporting and non-supporting evidence across existence, counting, and comparison tasks.}
        \label{fig:framework}
    \end{figure*}
    
    \subsection{Task-Specific Evidence Support}
    
    Let \(\hat{a}\) denote the extracted answer and \(n_c\) the number of covered instances of target class \(c\). We evaluate evidence support using the task-specific rules illustrated in Fig.~\ref{fig:framework}. For each sample, we set \(s=1\) when the grounding evidence supports \(\hat{a}\), and \(s=0\) for unsupported or invalid answers.
    
    \noindent\textbf{Existence.}
    For an existence question about class \(c\), a Yes answer requires at least one covered instance, while a No answer requires zero. We define evidence support as:
    \begin{equation}
    s =
    \begin{cases}
    1, & \hat{a}=\mathrm{Yes} \text{ and } n_c \geq 1,\\
    1, & \hat{a}=\mathrm{No} \text{ and } n_c = 0,\\
    0, & \text{otherwise}.
    \end{cases}
    \label{eq:existence_support}
    \end{equation}
    
    In Fig.~\ref{fig:framework}, covering one ship supports the Yes answer even though other ships are present. Full coverage of the target class is not required, since one instance is sufficient to support an existence claim. For No, the rule checks that no target instances are covered by the predicted regions. Target objects may still be present in the image without being covered.
    
    \noindent\textbf{Counting.}
    For a valid nonnegative integer answer \(\hat{a}\), the covered instance count must match the model's answer:
    \begin{equation}
    s =
    \begin{cases}
    1, & n_c=\hat{a},\\
    0, & \text{otherwise}.
    \end{cases}
    \label{eq:counting_support}
    \end{equation}
    
    In Fig.~\ref{fig:framework}, the image contains four airplanes, but the model answers three. Covering three instances supports this answer, whereas covering two does not. Covering all four also fails because the covered count exceeds the stated count. The rule checks agreement with the model's own answer, which can be supported even when incorrect. Evidence support and answer correctness are therefore evaluated separately.
    
    \noindent\textbf{Comparison.}
    Let \(n_A\) and \(n_B\) be the covered instance counts of classes \(A\) and \(B\). For questions asking whether \(A\) has more instances than \(B\), we define support as:
    \begin{equation}
    s =
    \begin{cases}
    1, & \hat{a}=\mathrm{Yes} \text{ and } n_A>n_B,\\
    1, & \hat{a}=\mathrm{No} \text{ and } n_A\leq n_B,\\
    0, & \text{otherwise}.
    \end{cases}
    \label{eq:comparison_support}
    \end{equation}
    
    For More-than questions, a No answer is supported even when both covered counts are zero. In Fig.~\ref{fig:framework}, covering three tennis courts and two basketball courts supports the Yes answer, whereas covering two of each does not. Neither class needs to be fully covered. Fewer-than questions use \(<\) for Yes and \(\geq\) for No; equality questions use \(=\) and \(\neq\), respectively.

    \begin{table*}[t]
    \centering
    \caption{Evaluation of seven vision--language models on G-RSVQA under the proposed protocol. All values are percentages; the best result in each column is bold.}
    \label{tab:main_results}
    \small
    \setlength{\tabcolsep}{5pt}
    \renewcommand{\arraystretch}{0.9}
    
    \begin{tabular*}{\textwidth}{
        @{\extracolsep{\fill}}
        c|*{7}{>{\fontsize{10}{12}\selectfont}c}@{}
    }
    \toprule
    
    \multicolumn{1}{@{}c!{\tikzmark{model-divider-top}\vrule width\arrayrulewidth}}{%
        \multirow{2}{*}[-0.8ex]{Model}}
    & \normalsize
      \multirow{2}{*}[-0.8ex]{\shortstack{Answer\\Accuracy}}
    & \multicolumn{3}{c}{\normalsize Localization}
    & \multicolumn{2}{c}{\normalsize Evidence Support}
    & \normalsize
      \multirow{2}{*}[-0.8ex]{\shortstack{Joint\\Success}} \\
    
    \cmidrule(lr){3-5}
    \cmidrule(lr){6-7}
    
    \multicolumn{1}{@{}c!{\tikzmark{model-divider-bottom}\vrule width\arrayrulewidth}}{}
    &
    & \normalsize Precision
    & \normalsize Recall
    & \normalsize F1-score
    & \normalsize Overall
    & \normalsize Correct Answers
    & \\
    
    \midrule
    
    GeoChat\cite{kuckreja2024geochat}
    & 60.11 & 32.87 & 1.95 & 3.68
    & 35.43 & 49.18 & 29.16 \\
    Qwen2.5-VL-7B\cite{wu2025qwen}
    & 65.50 & 41.05
    & 6.33 & 10.96
    & 58.94 & 92.82
    & 43.99 \\
    Kosmos-2\cite{peng2023kosmos}
    & 38.08 & 36.29 & 3.01 & 5.55
    & 51.20 & 79.90 & 47.12 \\
    GeoGround\cite{zhou2024geoground}
    & 67.38 & \textbf{52.27} & 4.40 & 8.11
    & 50.82 & 62.92 & 32.51 \\
    Ferret-7B\cite{you2024ferret}
    & 30.58 & 22.06 & 2.04 & 3.73
    & 30.34 & 76.80 & 27.58 \\
    InternVL3-8B\cite{zhu2025internvl3}
    & 76.92 & 17.47 & 3.14 & 5.32
    & 50.78 & 59.11 & 40.79 \\
    Qwen3-VL-8B  \cite{bai2025qwen3}
    & \textbf{77.47} & 31.63 & \textbf{10.05} & \textbf{15.26}
    & \textbf{80.11} & \textbf{94.35} & \textbf{63.76} \\
    
    \bottomrule
    \end{tabular*}%
    \begin{tikzpicture}[remember picture,overlay]
        \draw[line width=\arrayrulewidth]
            ([xshift=0.5\arrayrulewidth]pic cs:model-divider-top)
            --
            ([xshift=0.5\arrayrulewidth]pic cs:model-divider-bottom);
    \end{tikzpicture}\par
    
    \end{table*}

    \begin{table}[t]
    \centering
    \caption{Answer extraction accuracy (\%) on 400 VLM responses, evaluated against human reference annotations.}
    \label{tab:extractor_selection}
    \small
    \setlength{\tabcolsep}{2pt}
    \renewcommand{\arraystretch}{0.95}
    
    \begin{tabular*}{\columnwidth}{
        @{\extracolsep{\fill}}c|cccc@{}
    }
    \toprule
    Extractor & Existence & Counting & Comparison & Overall \\
    \midrule
    
    Qwen3-8B\cite{yang2025qwen3}
    & 99.32 & 83.02 & 88.89 & 95.75 \\
    
    \textbf{Qwen3-14B}\cite{yang2025qwen3}
    & \textbf{99.66}
    & \textbf{96.23}
    & \textbf{94.44}
    & \textbf{98.50} \\
    
    Mistral-NeMo-12B \cite{mistral2024nemo}
    & 96.25 & 67.92 & 92.59 & 92.00 \\
    
    \bottomrule
    \end{tabular*}
    \end{table}

    \subsection{Evaluation Metrics}
    
    We measure answer correctness using answer accuracy, with invalid answers counted as incorrect. Localization quality is measured by micro-averaged precision, recall, and F1-score using one-to-one matching between predicted and ground-truth boxes at IoU \(\geq 0.5\). Answer accuracy is reported on all 15,027 samples. For localization quality and evidence support, we exclude the 5,328 existence questions with No reference answers because their images contain no instances of the target class, leaving 9,699 samples for evaluation.
    
    Let \(\mathcal{E}\) denote the evaluation subset and \(s_i\) the evidence support indicator for sample \(i\). We set \(r_i=1\) if its extracted answer is correct and \(r_i=0\) otherwise. These indicators are used to compute evidence support and joint success.
    
    Evidence Support--Overall (\(\mathrm{ES}_{\mathrm{overall}}\)) measures the proportion of samples in \(\mathcal{E}\) whose grounding evidence supports the model's own answer, regardless of answer correctness:
    \begin{equation}
    \mathrm{ES}_{\mathrm{overall}}
    = \frac{\sum_{i\in\mathcal{E}} s_i}{|\mathcal{E}|}.
    \label{eq:es_overall}
    \end{equation}
    
    Evidence Support--Correct Answers (\(\mathrm{ES}_{\mathrm{correct}}\)) measures the proportion of correctly answered samples in \(\mathcal{E}\) whose grounding evidence supports the model's own answer:
    \begin{equation}
    \mathrm{ES}_{\mathrm{correct}}
    = \frac{\sum_{i\in\mathcal{E}} r_i s_i}
    {\sum_{i\in\mathcal{E}} r_i}.
    \label{eq:es_correct}
    \end{equation}
    
    Joint Success (\(\mathrm{JS}\)) measures the proportion of samples in \(\mathcal{E}\) with a correct answer and supporting grounding evidence:
    \begin{equation}
    \mathrm{JS}
    = \frac{\sum_{i\in\mathcal{E}} r_i s_i}{|\mathcal{E}|}.
    \label{eq:joint_success}
    \end{equation}

    \section{Experiment}
    \subsection{Experimental Setup}
    
    We evaluate the seven models in Table~\ref{tab:main_results} using publicly released weights, including GeoGround's task-specific LoRA weights, without additional training on G-RSVQA. Five models generate answers and grounding regions in one response; GeoChat and GeoGround use separate requests whose grounding prompts contain no previously generated answers. We adapt prompts to each model's interface, coordinate format and task. We compare three answer extractors on 400 VLM responses against human reference annotations. As shown in Table~\ref{tab:extractor_selection}, Qwen3-14B achieves the highest extraction accuracy of 98.50\% and is selected as our answer extractor.

    \subsection{Results and Analysis}
    
Table~\ref{tab:main_results} shows that answer accuracy and localization precision can not reflect evidence support. InternVL3-8B achieves the second-highest answer accuracy of 76.92\%, but its \(\mathrm{ES}_{\mathrm{overall}}\) is only 50.78\%. Qwen3-VL-8B also has a similar accuracy of 77.47\%, yet its \(\mathrm{ES}_{\mathrm{overall}}\) reaches 80.11\%. GeoGround achieves the highest precision of 52.27\%, but its \(\mathrm{ES}_{\mathrm{overall}}\) is 50.82\%. Qwen2.5-VL-7B shows how evaluating only correct answers can hide support failures: its support rate is 92.82\% among correct answers but 58.94\% across the evaluation subset. Its \(\mathrm{JS}\) of 43.99\% shows that fewer than half of this subset receive a correct answer and supporting evidence.

Table~\ref{tab:task} reveals a weakness in supporting exact counts. All seven models have lower \(\mathrm{ES}_{\mathrm{overall}}\) on counting than on existence. InternVL3-8B reaches 72.13\% on comparison but only 30.20\% on counting. Missing some objects may leave the relation between two class counts unchanged. A counting answer loses support whenever the covered count differs from the stated number. Strong comparison support therefore does not establish reliable support for exact counts. Qwen3-VL-8B illustrates why support and correctness must be distinguished: its counting \(\mathrm{ES}_{\mathrm{overall}}\) is 72.37\%, while its \(\mathrm{JS}\) is 50.69\%. This gap identifies answers whose grounding matches the stated number even though that number is incorrect.

\begin{table}[t]
\centering
\caption{Overall evidence support and joint success rates (\%) \textbf{by question type} on G-RSVQA.}
\label{tab:task}
\small
\setlength{\tabcolsep}{1.5pt}
\renewcommand{\arraystretch}{0.95}

\begin{tabular*}{\columnwidth}{
    @{\extracolsep{\fill}}c|*{6}{c}@{}
}
\toprule

\multicolumn{1}{
    @{}c!{\tikzmark{task-divider-top}\vrule width\arrayrulewidth}
}{%
    \multirow{2}{*}[-1.2ex]{Model}
}
& \multicolumn{2}{c}{Existence}
& \multicolumn{2}{c}{Counting}
& \multicolumn{2}{c}{Comparison} \\

\cmidrule(lr){2-3}
\cmidrule(lr){4-5}
\cmidrule(lr){6-7}

\multicolumn{1}{
    @{}c!{\tikzmark{task-divider-bottom}\vrule width\arrayrulewidth}
}{}
& \shortstack{ES\\Overall}
& \shortstack{Joint\\Success}
& \shortstack{ES\\Overall}
& \shortstack{Joint\\Success}
& \shortstack{ES\\Overall}
& \shortstack{Joint\\Success} \\

\midrule

GeoChat
& 62.93 & 60.69
& 8.87 & 6.22
& 46.44 & 26.90 \\

Qwen2.5-VL-7B
& \textbf{97.49} & 64.55
& 24.26 & 21.66
& 69.35 & 57.13 \\

Kosmos-2
& 90.84 & \textbf{90.84}
& 29.89 & 25.17
& 33.47 & 24.49 \\

GeoGround
& 88.33 & 70.13
& 28.89 & 6.69
& 37.55 & 26.76 \\

Ferret-7B
& 68.72 & 67.68
& 6.39 & 5.38
& 19.77 & 10.93 \\

InternVL3-8B
& 63.60 & 54.60
& 30.20 & 23.16
& \textbf{72.13} & 54.81 \\

Qwen3-VL-8B
& 95.99 & 83.18
& \textbf{72.37} & \textbf{50.69}
& 71.44 & \textbf{60.28} \\

\bottomrule
\end{tabular*}%
\begin{tikzpicture}[remember picture,overlay]
    \draw[line width=\arrayrulewidth]
        ([xshift=0.5\arrayrulewidth]pic cs:task-divider-top)
        --
        ([xshift=0.5\arrayrulewidth]pic cs:task-divider-bottom);
\end{tikzpicture}\par

\end{table}

    \section{Conclusion}

We proposed AESP to evaluate answer-grounding consistency in grounded RS-VQA and constructed the G-RSVQA benchmark with 15,027 samples. Experiments on seven grounded vision--language models show that answer correctness and localization quality do not necessarily imply supporting grounding evidence, with counting showing particularly weak support. These results highlight the importance of explicitly evaluating whether grounding evidence supports a model's own answer.


    \bibliographystyle{IEEEbib}
    \bibliography{refs}
    
    \end{document}